\pdfoutput=1
\documentclass[11pt]{article}

\usepackage[]{acl}

\usepackage{times}
\usepackage{latexsym}

\usepackage[T1]{fontenc}
\usepackage[utf8]{inputenc}

\usepackage{microtype}

\usepackage{inconsolata}

\usepackage{url}
\usepackage[T1]{fontenc}
\usepackage{multirow}
\usepackage{listings}
\usepackage{graphicx}

\graphicspath{ {./figures/} }

\makeatletter
\def\thanks#1{\protected@xdef\@thanks{\@thanks
        \protect\footnotetext{#1}}}
\makeatother

\title{FiNER-ORD: Financial Named Entity Recognition Open Research Dataset}

\author{\hypersetup{linkcolor=black} Agam Shah, Abhinav Gullapalli, Ruchit Vithani, Michael Galarnyk, Sudheer Chava\\
Georgia Institute of Technology
\thanks{Correspondence to Agam Shah \textcolor{darkblue}{{\{\href{mailto:ashah482@gatech.edu}{ashah482@gatech.edu}\}}}}}

\begin{document}
\maketitle
\begin{abstract}

Over the last two decades, the development of the CoNLL-2003 named entity recognition (NER) dataset has helped enhance the capabilities of deep learning and natural language processing (NLP). The finance domain, characterized by its unique semantic and lexical variations for the same entities, presents specific challenges to the NER task; thus, a domain-specific customized dataset is crucial for advancing research in this field. In our work, we develop the first high-quality English Financial NER Open Research Dataset (FiNER-ORD). We benchmark multiple pre-trained language models (PLMs) and large-language models (LLMs) on FiNER-ORD. We believe our proposed FiNER-ORD dataset will open future opportunities to use FiNER-ORD as a benchmark for financial domain-specific NER and NLP tasks. Our dataset, models, and code are publicly available on \href{https://github.com/gtfintechlab/FiNER-ORD}{GitHub} and \href{https://huggingface.co/datasets/gtfintechlab/finer-ord}{Hugging Face} under CC BY-NC 4.0 license. 
\end{abstract}

\section{Introduction}
The growth of technology and the web over the last several decades has led to a rapid increase in the generation of text data, especially in the financial domain. The abundance of financial texts, primarily in the form of news and regulatory filings, presents a valuable resource for analysts, researchers, and even individual investors to extract relevant information. With the given abundance of text data, manually extracting relevant information is impossible and unsustainable for large-scale text datasets used for rapid downstream tasks. However, NLP techniques can help with automating the information retrieval process. Named entity recognition (NER) is one such NLP task that serves as an important first step to identify named entities, such as persons, organizations, and locations, and efficiently use available text data to generate a knowledge graph.

While numerous studies have constructed annotated datasets \citep{sang2003introduction, derczynski2017results, AB2-MKJJ2R-2013} and developed NER models \citep{li2019dice, yamada2020luke, wang2020automated, wang2021improving} for generic texts, the financial domain presents unique challenges that necessitate the use of domain-specific text. The current annotated dataset based on Credit Risk Agreements \cite{alvarado2015domain} for financial NER has significant shortcomings, limiting the use of deep learning models for financial NER. We elaborate on these shortcomings in Section \ref{sec:dataset_comparison}.

In addressing these challenges within the scope of data customization and ethics-conscious evaluation, our work contributes significantly to the field by establishing the largest Financial Named Entity Recognition Open Research Dataset (FiNER-ORD). Our objective in this context is to create an evaluation dataset and evaluate existing generalist, non-customized models. We evaluate pre-trained language models (PLMs) and zero-shot large-language models (LLMs) on the FiNER-ORD dataset.

\begin{figure}[ht]
    \centering
        \includegraphics[width=0.45\textwidth]{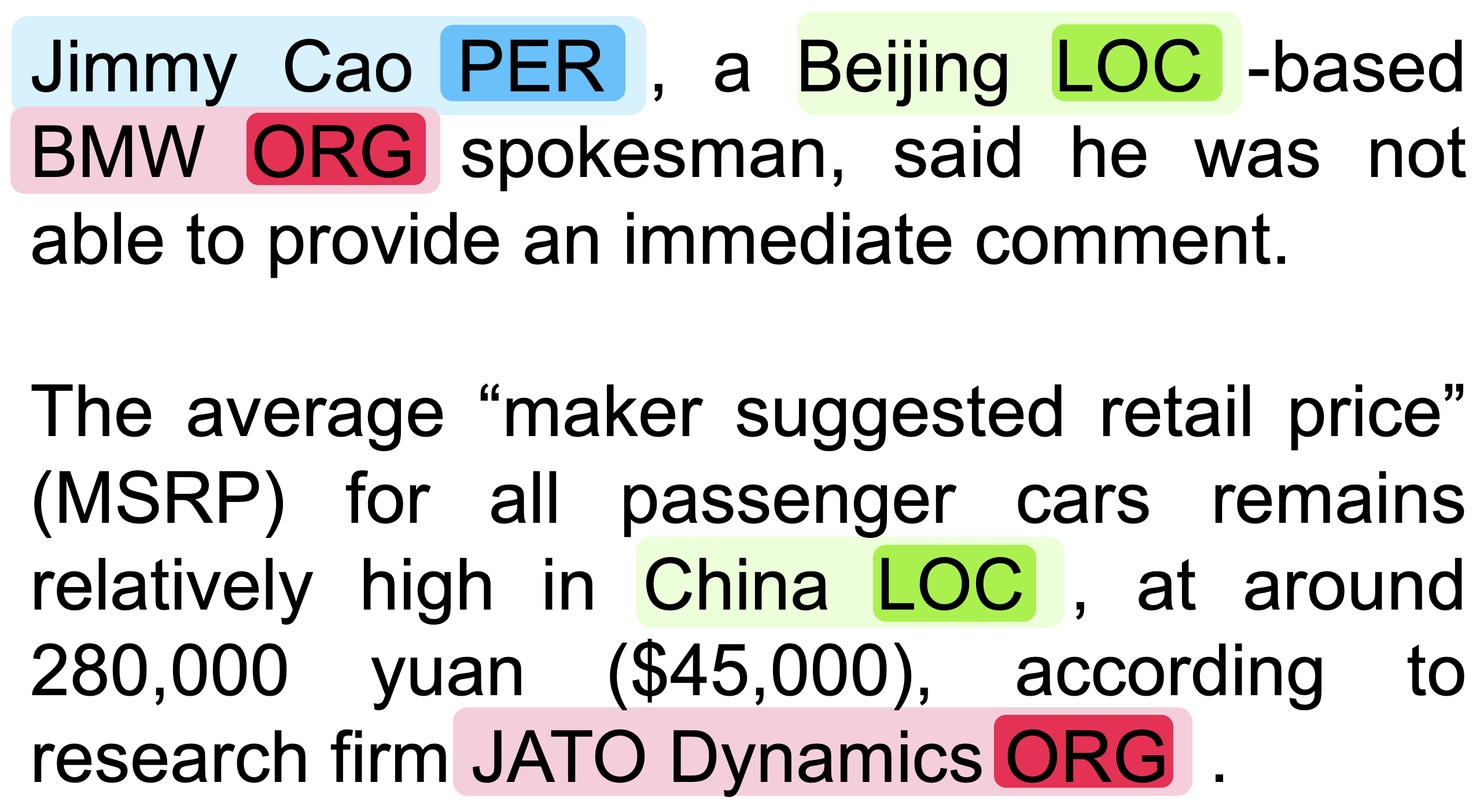}
    \caption{Representative example of annotation in FiNER-ORD.}
    \label{fig:nerexample}
\end{figure}

\section{Related Work}

\paragraph{Named Entity Recognition Datasets} 
Commonly used general NER datasets include the CoNLL-2003 \citep{sang2003introduction}, OntoNotes5.0 \citep{AB2-MKJJ2R-2013}, WNUT 2017 \citep{derczynski-etal-2017-results}, and FewNERD \citep{ding-etal-2021-nerd}. FiNER-139 from \citet{loukas-etal-2022-finer} is an entity recognition dataset for numerical financial data. Like our work, they also don't use numerical entities because they can be easily extracted using regular expressions and are the easiest entity types to recognize. There has also been work in recognizing entities in invoices, business forms, and emails \citep{info10080248}. CRA NER dataset from \citet{alvarado2015domain} was the first attempt to create a financial NER dataset similar to FiNER-ORD, but in our assessment presented in Section \ref{sec:dataset_comparison}, it is skewed and thus limited in usefulness.

\paragraph{Applications for Information Retrieval in Finance} 
Sentiment analysis of news aids event ranking based on market factors, aiding price prediction \citep{feng2021fineventranking}. Analyzing tweets and headlines automates trading, price movement, and risk forecasts \citep{sawhney2021hyperbolic}. Extracting claims from analysts' reports improves volatility forecasting on release and earning dates \cite{shah2024numerical}. Real-time web tools extract operating segments, aiding performance analysis \citep{ma2020opsegspot}. Transformer models fine-tuned for price-change data extraction measure inflation exposure \cite{chava2022measuring}. Financial knowledge graphs from news data enhance algorithmic trading \citep{cheng2020finkgtrading}.

\section{Dataset}

\subsection{FiNER-ORD}
\label{sec:news_data}
FiNER-ORD consists of a manually annotated dataset of financial news articles (in English) collected from webz.io\footnote{\url{https://webz.io/free-datasets/financial-news-articles/}}. In total, there are 47,851 news articles available for download in this dataset at the time of writing this paper. 

\paragraph{Sampling and Manual Annotation}
\label{sec:samp&manual_annot}
 For the manual annotation of named entities in financial news, we randomly sampled 220 documents from the entire set of news articles. We observed that some articles were empty in our sample, so after filtering the empty documents, we were left with a total of 201 articles. We use the open-source Doccano\footnote{\url{https://github.com/doccano/doccano}} annotation tool (see Appendix \ref{app:annotation_tool} for more details) to ingest the raw dataset and manually label person (PER), location (LOC), and organization (ORG) entities. Figure~\ref{fig:nerexample} shows an example of manually annotated named entities in FiNER-ORD. Labeling was performed independently by two different annotators, following the predefined annotation guide presented in Appendix \ref{app:annotation_guide}, to reduce potential labeling bias. Annotator labeling was compared and validated for consistency. We provide details about annotator background in Appendix \ref{app:annotator_background}, annotator agreement in Appendix \ref{app:annotator_aggrement}, and post-processing in Appendix \ref{app:post_process}.

\paragraph{Dataset Statistics}

 Following the manual annotation and post-processing procedures, each token is labeled as one of 4 broad entity types: PER, LOC, ORG, and O. As discussed in Appendix \ref{app:post_process} each of the PER, LOC, and ORG classes are further segmented with the suffixes \_B (denoting beginning token of a span) and \_I (denoting intermediate token of a multi-token span), respectively. As discussed earlier, we manually annotate the train, validation, and test splits of the 201 articles in FiNER-ORD. The descriptive statistics on the resulting FiNER-ORD are available in Table~\ref{tb:dataset_finerord}. 

\begin{table}
\centering
\footnotesize
\begin{tabular}{lccc}
\hline
\textbf{FiNER-ORD} & \textbf{Train} & \textbf{Validation} & \textbf{Test}\\
\hline
Articles & 135 & 24 & 42\\
Sentences & 3,262 & 402 & 1,075\\
Tokens & 80,531 & 10,233 & 25,957\\
PER & 821 & 138 & 284\\
LOC & 966 & 193 & 300\\
ORG & 2,026 & 274 & 544\\
\hline
\end{tabular}
\caption{Number of articles, sentences, tokens, and entities (person, location, organization) in the train, validation, and test splits of FiNER-ORD.}
\label{tb:dataset_finerord}
\end{table}

\subsection{Comparison of Datasets}
\label{sec:dataset_comparison}

\paragraph{Credit Risk Agreements (CRA)}
Table \ref{tb:dataset_comparison} compares our proposed dataset (FiNER-ORD) with the CRA NER dataset \citep{alvarado2015domain}. The CRA dataset attempts to provide a domain-specific dataset in CoNLL format using manual annotation on 8 English documents from the U.S. Security and Exchange Commission (SEC) filings \citep{bird2009nltk}. However, the annotation methodology for the CRA NER dataset "automatically tagged all instances of the tokens \textit{lender} and \textit{borrower} as being of entity type PER" \citep{alvarado2015domain}. This approach is problematic because of the resulting skewed distribution of entity types in the dataset, leading to confounded results. Our analysis of the CRA NER dataset showed that in FIN3 (CRA test data split), instances of the tokens \textit{lender} and \textit{borrower} represented 83.05\% of all PER tokens and 44.95\% of all tokens labeled as PER, ORG, MISC, or LOC. Similarly, in FIN5 (CRA train data split), instances of the tokens \textit{lender} and \textit{borrower} represented 90.04\% of all PER tokens and 46.08\% of all tokens labeled as PER, ORG, MISC, or LOC.

Thus, we believe the CRA dataset is not a high-quality benchmark for specialized NLP tasks in the financial domain, motivating us to create a new high-quality financial domain-specific NER dataset. The dataset comparison statistics are available in Table~\ref{tb:dataset_comparison}. We also perform a transfer learning study presented in Section \ref{app:transfer_learning} in order to further emphasize the importance of FiNER-ORD in comparison to the CRA NER dataset.

\paragraph{CoNLL-2003}
The CoNLL NER dataset \citep{sang2003introduction} was created with manual annotation on Reuters generic news stories. In general, financial texts differ from general texts and contain a higher ratio of organization tokens and entities compared to person and location tokens and entities. This can be important for financial news network applications that need to capture the interdependence of companies and the complexity of modern financial markets. In the FiNER-ORD dataset, the ratio of organizations (ORG) to location (LOC) to person (PER) entities is 2.29:1.17:1. This distribution contrasts with the CRA and CoNLL datasets, which exhibit ORG:LOC:PER ratios of 0.31:0.22:1 and 0.93:1.06:1, respectively. 

We also analyze the average length of each entity-type in terms of the number of tokens per entity. In the order of FiNER-ORD, CRA, and CoNLL, we see the average number of tokens for a PER entity to be approximately 1.66, 1.06, and 1.69, respectively. Similarly, for LOC entities we find an average length of approximately 1.34, 2.09, and 1.50, respectively. For ORG entities, we find an average length of 1.72, 1.69, 1.18, respectively. These findings show that for PER entities, the average length is similar in FiNER-ORD and CoNLL. However, the skewed distribution of PER entities with the problematic labeling of \textit{lender} and \textit{borrower} as PER in CRA leads to an average length value close to 1 for PER entities in CRA, underscoring the finance-specific utility of FiNER-ORD over CRA. 

Similarly, for LOC entities, the average length values are similar in FiNER-ORD and CoNLL. However, the finance-specific usefulness of FiNER-ORD over CoNLL is highlighted by the average length values for ORG entities. We see similar average lengths for ORG entities in FiNER-ORD and CRA, both of which are financial NER datasets, but the average length of ORG entities in CoNLL is much smaller. Additionally, the comparison between the ratios of ORG\_B and ORG\_I tokens for FiNER-ORD (1.4:1) and CoNLL (5.6:1) suggests that in financial texts, organization entities are more likely to span over multiple tokens. Having a higher percentage of useful ORG entities and tokens can also enhance applications like measuring correlations in financial news networks and market movements \citep{wan2021sentiment}. This highlights the importance of creating high-quality datasets specifically for the financial domain.

\begin{table}
\centering
\footnotesize
\begin{tabular}{lccc}
\hline
\textbf{Dataset} & \textbf{FiNER-ORD} & \textbf{CRA} & \textbf{CoNLL}\\
\hline
Articles & 201 & 8 & 1,393\\
Sentences & 4,739 & 1,473 & 22,137\\
Tokens & 116,721 & 54,262 & 301,418\\
PER & 1,243 & 962 & 10,059\\
LOC & 1,459 & 208 & 10,645\\
ORG & 2,844 & 295 & 9,323\\
\hline
\end{tabular}
\caption{Comparison of our FiNER-ORD with Credit Risk Agreements and CoNLL-2003 English NER datasets in terms of the number of articles, sentences, tokens, and entities (person, location, organization).}
\label{tb:dataset_comparison}
\end{table}

\begin{table*}[ht]
\centering
\footnotesize
\begin{tabular}{lccccc}
  \hline
  \textbf{Model} & \textbf{Train Split} & \textbf{PER} & \textbf{LOC} & \textbf{ORG} & \textbf{Weighted Average} \\ 
  \hline
     \multicolumn{6}{c}{\textbf{Panel A: FineTuning with PLMs}} \\ 
    \hline
  BERT-base-cased & FiNER-ORD & 0.8811 (0.0192) & 0.6820 (0.0239) & 0.6013 (0.0490) & 0.6931 (0.0327) \\ 
  FinBERT-base-cased & FiNER-ORD & 0.7456 (0.0254) & 0.6836 (0.0238) & 0.6002 (0.0288) & 0.6589 (0.0231)\\  
  RoBERTa-base & FiNER-ORD & 0.9050 0.0076) & 0.7154 (0.0608) & 0.6304 (0.0878) & 0.7220 (0.0585)\\ 
  \hline
  BERT-large-cased & FiNER-ORD & 0.8954 (0.0090) & 0.7289 (0.0467) & 0.6272 (0.0145) & 0.7216 (0.0039)\\ 
  RoBERTa-large & FiNER-ORD  & \textbf{0.9263} (0.0025) & \textbf{0.7717} (0.0152) & \textbf{0.6769} (0.0130) & \textbf{0.7648} (0.0057)\\
   \hline
   \multicolumn{6}{c}{\textbf{Panel B: Zero-Shot with Generative LLMs}} \\ 
    \hline
   Llama-3.1-70B-Turbo & Zero-Shot & 0.6706 (0.0033) & 0.5919 (0.0027) & 0.4981 (0.0038) & 0.5661 (0.0004)\\
   Llama-3.1-405B-Turbo & Zero-Shot & 0.7413 (0.0020) & 0.6587 (0.0017) & 0.5751 (0.0024) & 0.6389 (0.0016)\\
   GPT-4o & Zero-Shot & 0.8004 (0.0065) & 0.6651 (0.0073) & 0.6036 (0.0111) & 0.6692 (0.0057)\\
    \hline
    \multicolumn{6}{c}{\textbf{Panel C: Transfer Learning Ablation}} \\ 
    \hline
      RoBERTa-large & CRA  & 0.5918 (0.0868) & 0.0145 (0.0020) & 0.0411 (0.0174) & 0.1730 (0.0262)\\
      RoBERTa-large & CoNLL  & 0.8707 (0.0171) & 0.7640 (0.0278) & 0.5668 (0.0678) & 0.6954 (0.0278)\\
    \hline
\end{tabular}

\caption{Performance comparison of various models tested on the FiNER-ORD test split. Panel A contains results for various models fine-tuned on the FiNER-ORD train sample. Panel B contains results for zero-shot Generative LLMs. Panel C contains results for transfer learning experiments where the RoBERTa-large model is fine-tuned on the Credit Risk Agreements (CRA) NER dataset and CoNLL-2003 dataset. All values are weighted F1 scores. An average of 3 seeds was used for all models. The standard deviation of the F1 scores is reported in parentheses.}
\label{tb:master_performance}
\end{table*}

\section{Models}

\subsection{PLMs}
We benchmark FiNER-ORD with several base and large transformer-based models. For the base model category, we use BERT \citep{bert}, FinBERT \citep{finbert_yang2020}, and RoBERTa \citep{roberta}. For the large model category, we use BERT-large \citep{bert} and RoBERTa-large \citep{roberta}. We do not pre-train these models before fine-tuning them. Further details about fine-tuning is provided in Appendix \ref{app:best_hyperparam_configs}.

\subsection{LLMs}
\label{sec:llm-exps}
To benchmark the performance of current SOTA generative LLMs, we measure the zero-shot performance on three train test splits for the "gpt-4o-2024-08-06"\footnote{\url{https://platform.openai.com/docs/guides/gpt}}, "Meta-Llama-3.1-405B-Instruct-Turbo", and "Meta-Llama-3.1-70B-Instruct-Turbo" models with 0.0 temperature and 1000 max tokens for the output. All API calls were made on August 8, 2024 or August 9, 2024. The zero-shot prompt can be found in Appendix \ref{app:zero_shot_prompt}.

\section{Results and Analysis}
In this section, we evaluate and benchmark different NLP models on the NER task with FiNER-ORD. For all models, we consistently use the same train, validation, and test splits in FiNER-ORD at the sentence-level.

For each PLM, we use three different seeds for the three runs of each model. We run each LLM three times with the same settings discussed in Section \ref{sec:llm-exps}. We report average weighted F1 scores for the best hyper-parameter configuration of each model in Table \ref{tb:master_performance}. 

Although zero-shot GPT-4o outperforms zero-shot the Llama models, fine-tuned PLMs outperform both zero-shot LLMs across all entity label categories. This finding aligns with the survey by \citet{pikuliak_chatgpt_survey}, which finds that zero-shot ChatGPT fails to outperform fine-tuned models on more than 77\% of NLP tasks. RoBERTa-base achieves best weighted F1 score overall and for all entity label categories when compared to all models. A future ablation study of these language models would provide key insights such as why RoBERTa-base, a model more generally trained with the masked language modeling (MLM) objective, outperforms (when considering our financial NER evaluation) FinBERT-base, a model trained on financial sentiment classification tasks and widely used for various financial domain-specific NLP tasks. Applying FiNER-ORD as a financial-domain NER benchmark dataset for such studies can further improve customized financial domain-specific language models.

\subsection{Transfer Learning Ablation}
\label{app:transfer_learning}
This study evaluates the transfer learning capabilities of the best-performing model, RoBERTa-large, originally trained and tested on the FiNER-ORD dataset. Specifically, we investigate its performance when fine-tuned on the CRA NER dataset and CoNLL dataset, followed by testing on the test split of the FiNER-ORD dataset. The outcomes of these transfer learning experiments are detailed in Panel C of Table \ref{tb:master_performance}. We also evaluate an existing model trained on the CONNL 2003 dataset and test it on FiNER-ORD in Appendix \ref{app:train_CoNLL}. These results underscore the significance of the FiNER-ORD dataset in enhancing model performance relative to the existing CRA and CoNLL NER datasets.

\section{Conclusion}
We present the first high-quality, manually annotated financial NER dataset, FiNER-ORD, which was generated from open-source financial news articles. We demonstrate the importance of our FiNER-ORD dataset compared to the existing CoNLL-2003 and financial CRA NER datasets. To evaluate the proposed dataset, we benchmark various configurations of PLMs and LLMs on FiNER-ORD. The performance analysis shows that RoBERTa-base outperforms all tested models (including FinBERT-base) overall, and GPT-4o, Llama-3.1-405B, and Llama-3.1-70B underperform the tested PLMs over all entity label categories. Furthermore, our proposed financial domain-specific dataset and performance analysis together present opportunities to further explore improvements for customized financial domain-specific language models.

\section*{Ethics Statement}
All language models used, under their respective license categories, are publicly available for our experimental purposes. With regards to ethical considerations for the environmental impact of our experiments, we only fine-tune the benchmarked PLMs because pre-training PLMs are known to have a large carbon footprint. We acknowledge that our dataset is constructed using English news articles and thus biased in terms of language representation and inclusion. The news articles do not contain offensive or discriminatory content. The dataset license and copyright exceptions are stated in Appendix \ref{app:copyright_restrictions}.

\section*{Limitations}
We acknowledge a geographic bias in the dataset as FiNER-ORD only includes English financial news articles. In the future, other forms and languages of financial texts could be manually annotated to expand FiNER-ORD. Additionally, more label classes such as `product', `miscellaneous', and forms of relationships could be annotated for future downstream applications such as knowledge graph creation. We do not present a new SOTA model architecture for financial NER and we do not include or discuss finance domain-specific LLMs
like BloombergGPT \cite{wu2023Bloomberg} in our work as we have no way of accessing it. Our primary focus of this work is to present the first high-quality financial NER open research dataset and benchmark our FiNER-ORD with multiple PLMs and LLMs to evaluate the performance of these models on the finance domain-specific NER task.

\paragraph{Form 10-K Filings} 
When considering data sources for financial NER, we found that SEC Form 10-K filings, filed annually by public companies in the United States, have certain limitations. The 10-K provides details about a company's business, risks, and operating and financial results for each fiscal year. For the NER task, however, we observe there is a noticeable bias towards organization entity tokens associated with the company filing the 10-K. As a result, there is a reduced diversity of named entities, particularly concerning other organizations. Furthermore, the average length of 10-K filings has been growing significantly every year with at least 60,000 words between 2022 and 2023, a significant increase from around 30,000 words in 2000 and approximately 42,000 words in 2013\footnote{\url{https://www.wsj.com/articles/the-109-894-word-annual-report-1433203762}}. In contrast, financial news articles are shorter and can be used to better capture the interdependence of companies and the complexity of modern financial markets. Given this, we decided that constructing a high-quality, manually annotated dataset would be more feasible and information-diverse with the open-source news articles we annotate in FiNER-ORD as opposed to 10-K filings.

\section*{Acknowledgements}

We appreciate the generous support of Azure credits from Microsoft made available for this research via the Georgia Institute of Technology Cloud Hub. We would like to thank Manan Jagani, Darsh Rank, Visaj Shah, Nitin Jain, Roy Gabriel, Olaolu Dada, and Gabriel Shafiq for their contribution to the project in the initial stage.

\bibliography{custom, anthology}

\appendix

\section{Copyright Exceptions}
\label{app:copyright_restrictions}
The dataset will be made publicly available on Hugging Face under the non commercial CC BY-NC 4.0 license. The individual articles comprising the dataset can be considered exempt from copyright for non-commercial research. Exceptions to copyright in the United Kingdom are regularly updated at gov.uk\footnote{\url{https://www.gov.uk/guidance/exceptions-to-copyright}}. The EU Directive on Copyright in the Digital Single Market\footnote{\url{https://eur-lex.europa.eu/eli/dir/2019/790/oj}} provides exceptions for reproductions made by research organizations.

\section{Annotator Background}
\label{app:annotator_background}
Labeling was performed independently by two different annotators. Annotator 1 is a Doctoral Researcher from India working with NLP and Computational Finance. Annotator 2 is a Masters student from the United States pursuing a Computer Science degree with Machine Learning specialization. Both annotators were male researchers and read English financial news articles daily from several outlets and sources. Additionally, both annotators are authors, so the annotators were not hired or paid for annotations.

\section{Annotator Agreement}
\label{app:annotator_aggrement}
Both annotators had a manual annotation agreement of approximately 96.85\% of the 5546 entities across the PER, LOC, ORG entity classes and train, validation, test splits of FiNER-ORD. The annotation guide was referenced to resolve disagreements between the manual labeling of both annotators. The annotation guide was developed iteratively during the annotation process. Online resources were consulted if the annotation guide did not address a specific disagreement, and the annotation guide was updated accordingly afterwards.

\section{Post-Processing}
\label{app:post_process}
To correct potential errors in manual annotations, we run a custom post-processing script that performs the following four tasks: (1) remove trailing spaces from annotated entities, (2) extend token-level borders to non-space characters to change an erroneous span to the correct span, (3) clean entity suffixes with techniques such as removing an apostrophe followed by the letter s (\textit{'s}) from entity suffixes, (4) tokenize text with Stanza \citep{qi-etal-2020-stanza} and add positional information for labeled entities by splitting multi-token spans into separate tokens, assigning \_B as the label suffix for the first separated token in the multi-token span, and assigning \_I as the label suffix for the remaining separated tokens in the multi-token span. We note that after the post-processing script is run, all tokens which are not annotated with one of PER\_B, PER\_I, LOC\_B, LOC\_I, ORG\_B, ORG\_I are assigned the label O, denoting "other" type of token not belonging to the person, location, organization classes.

\begin{table*}[ht]
\centering
\footnotesize
\begin{tabular}{p{0.22\textwidth}p{0.2\textwidth}p{0.2\textwidth}p{0.2\textwidth}}
\hline
\textbf{All Entities} & \textbf{PER}& \textbf{LOC}& \textbf{ORG}\\
\hline
   China-155 &               Obama-26 &        China-154 &          GM-52 \\
    U.S.-69 &              Abbott-19 &         U.S.-69 &     Reuters-47 \\
       GM-52 &             Clinton-18 &        Greece-49 &    Facebook-35 \\
   Greece-49 &                Bush-14 &            US-46 &      Nikkei-34 \\
  Reuters-47 &            Turnbull-14 &            UK-37 &         Fed-32 \\
       US-46 &          Varoufakis-11 &     Australia-32 &       Apple-31 \\
       UK-37 &              Jaitley-9 &        London-28 &        Ford-24 \\
 Facebook-35 &              Shorten-9 &         Japan-24 &   Stratasys-23 \\
   Nikkei-34 &              McPhail-9 &      New York-23 &         LSE-21 \\
Australia-32 &          Tony Abbott-8 &        Europe-22 &       House-20 \\
      Fed-32 &     Malcolm Turnbull-8 &        Sydney-21 &  UK Markets-19 \\
    Apple-31 &               Andrew-8 &        Taiwan-19 &    Unilever-19 \\
   London-28 &               Bishop-8 &       Bahrain-18 & Motley Fool-18 \\
    Obama-26 & John - Erik Koslosky-8 &         Kenya-18 &         SEC-18 \\
    Japan-24 &                Glatt-7 &         India-17 &          FT-18 \\
\hline
\end{tabular}
\caption{Most common entities in FiNER-ORD.}
\label{tb:most_common_entities}
\end{table*}

\section{Annotation Guide}
\label{app:annotation_guide}
The manual annotation process to create FiNER-ORD consisted of ingesting the financial news articles in Doccano. Each news article is available in the form of a JSON document with various metadata information including the source of the article, publication date, author of the article, and the title of the article. Entities of the type \textit{person} (PER), \textit{organization} (ORG), and \textit{location} (LOC) were identified according to the rules described below. Some well-known names for these entities were obvious while others were confirmed by researching the names to identify the correct entity type.

\subsection{Person Entities}
PER entities were identified by their first name and/or last name. In the examples below, bold spans represent a single person entity. In the case where a person was identified by their first and last name, the entire name was labeled as PER and the post-processing script tagged the first name as PER\_B and the last name as PER\_I. Words like \textit{President}, \textit{Ms}, and \textit{CEO} were not labeled as part of the PER entity but help indicate a PER entity. In a context indicating possession with \textit{'s}, the name until the \textit{'s} was tagged.
\begin{itemize}
    \item President \textbf{Obama}
    \item CEO \textbf{Phyllis Wakiaga}
    \item Ms \textbf{Wakiaga}
    \item \textbf{Bill Clinton}'s
\end{itemize}

\subsection{Location Entities}
LOC entities primarily consisted of names of continents, countries, states, cities, and addresses. In the examples below, bolded spans represent spans comprising LOC entities. Commas in addresses are not included tagged LOC entities. In such cases for tagging addresses, each complete span delimited by a comma was tagged as a LOC entity. In a context indicating possession with \textit{'s}, the name until the \textit{'s} was tagged. In the case where a location was identified by multiple tokens delimited by a space, the entire name was labeled as LOC and the post-processing script tagged the first name as LOC\_B and the last name as LOC\_I. Words such as "Kenyan" were treated as adjectives and thus not labeled as a LOC entity. When discussing a lawmaker's political and location affiliation, examples such as \textit{R-Texas} denoting "Republican from Texas" are encountered, in which only the location name such as \textit{Texas} is tagged.
\begin{itemize}
    \item \textbf{Asia}
    \item \textbf{US}
    \item \textbf{India}
    \item \textbf{United States}
    \item \textbf{Beijing}
    \item \textbf{New York}
    \item \textbf{Redwood City}, \textbf{California}
    \item \textbf{Kenya}'s
    \item \textbf{Mombasa Road}
    \item R-\textbf{Texas}
\end{itemize}

\subsection{Organization Entities}
ORG entities consist of examples such as company names, news agencies, government entities, and abbreviations such as stock exchange names and company stock tickers. Punctuation marks such as hyphens are included when tagging an ORG entity. As designated, \textit{.com} is included in the identified company's name. In a context indicating possession with \textit{'s}, the name until the \textit{'s} was tagged.
\begin{itemize}
    \item \textbf{Wal-Mart}
    \item \textbf{China Resources SZITIC Trust Co Ltd}
    \item \textbf{The Wall Street Journal}
    \item \textbf{Atlanta Federal Reserve} 
    \item \textbf{Morgan Stanley}'s
    \item \textbf{Delta Air Lines}
    \item \textbf{DAL}
    \item \textbf{NYSE}
    \item \textbf{Amazon.com}
\end{itemize}

\subsection{Annotation Edge Cases}
There were a few edge cases in the annotation process. For example, labeling entities when location (LOC) is part of the organization (ORG) entity is a common problem in finance. The phrase "Google India" has "India" which is a location, but it is labeled as an organization in our framework. This is because our process does not permit overlapping entity labels. Another such example is "New York Stock Exchange" which we annotated entirely as an ORG entity, despite models often predicting “New York” as LOC. Therefore, the correct labels are ORG\_B for “New” and ORG\_I for “York,” whereas models might label them as LOC\_B and LOC\_I, respectively. Commonly used words like "the" which are often tagged as O (Other) may sometimes be present in the name of an organization, such as "The Wall Street Journal". In such cases, we have tagged "The" to be included as part of an ORG entity. In the specific example for "The Wall Street Journal", as shown in Appendix \ref{app:annotation_guide}, we have tagged "The" as ORG\_B and the remaining tokens "Wall", "Street", "Journal" as ORG\_I tokens. For entities which may represent both an organization and a product, such as "Google", we have only tagged such entities as ORG because our dataset currently does not provide manual annotations for product entities.

\section{Most Common Entities}

 Table \ref{tb:most_common_entities} shows the most common entities within FiNER-ORD. Since the dataset is comprised of financial news articles from July to October 2015, the dataset has a bias towards the news of that time period. An advantage of FiNER-ORD is that it has a unique heterogeneity due to it being composed of English language financial articles published by institutions from around the world unlike the CRA dataset which are from United States Securities and Exchange Commission documents. The most common LOC and ORG entities reflect the global news article source.

\section{Doccano Annotation Tool}
\label{app:annotation_tool}
\begin{figure}[h]
    \centering
        \includegraphics[width=0.45\textwidth]{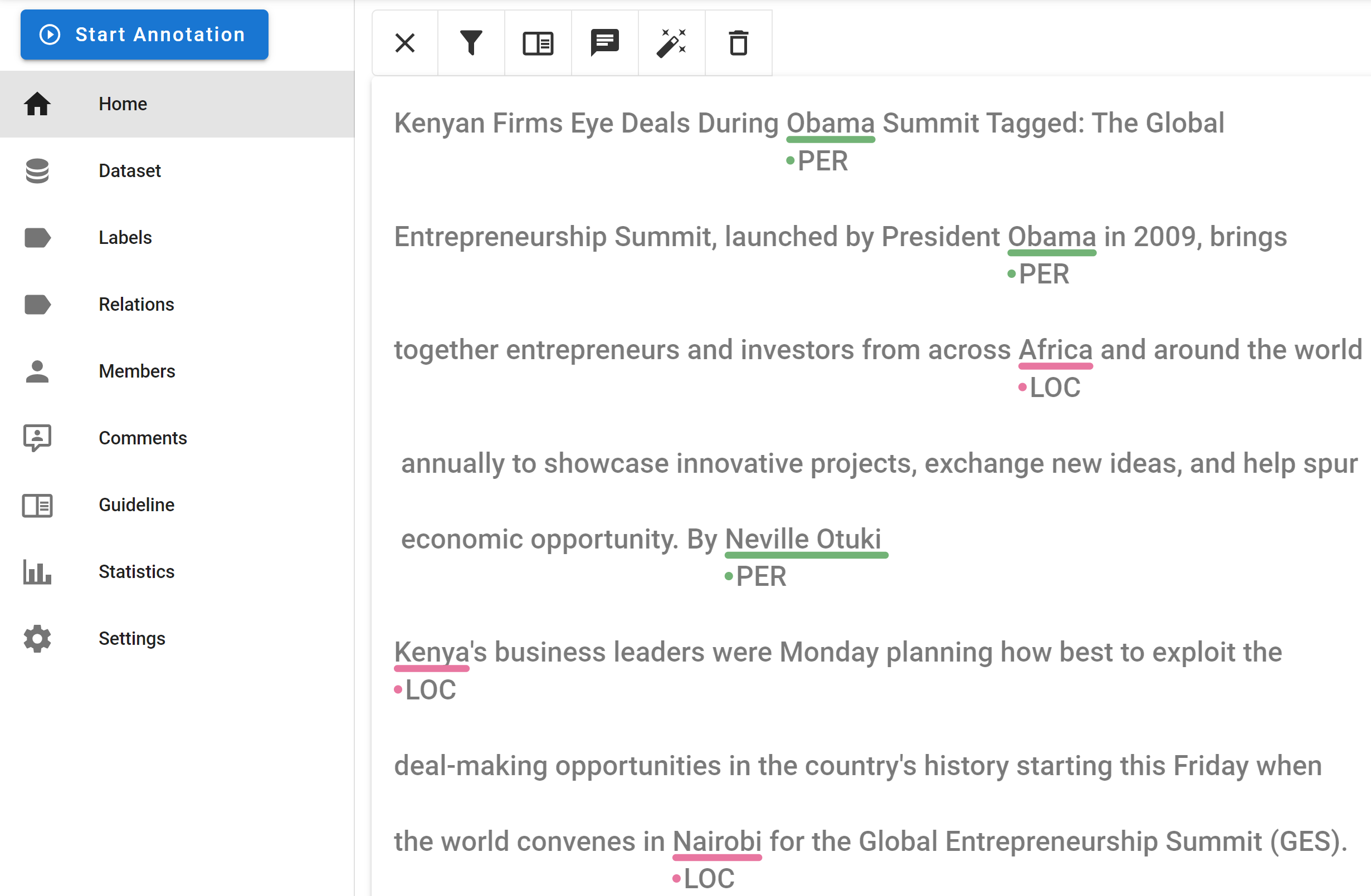}
    \caption{Screenshot of an article in FiNER-ORD manually annotated with the open-source Doccano annotation tool.}
    \label{fig:doccano_ss}
\end{figure}

All manual annotation of FiNER-ORD was completed using the open-source Doccano\footnote{\url{https://github.com/doccano/doccano}} annotation tool. Figure \ref{fig:doccano_ss} demonstrates the use of Doccano to manually annotate FiNER-ORD. The output from Doccano contains span-level label information. This information is in the form of a list of lists containing information on the start character, end character, and label of each entity annotated by the manual annotator.

\begin{table}
\centering
\footnotesize
\begin{tabular}{lcc}

\hline
\textbf{Model} & \textbf{learning rate} & \textbf{batch size}\\
\hline
BERT-base-cased & 1e-5  & 16\\
FinBERT-base-cased & 1e-5 & 8\\
RoBERTa-base & 1e-5 &  8\\
BERT-large-cased & 1e-5 & 8\\
RoBERTa-large & 1e-5 &  8\\
\hline
\end{tabular}
\caption{Best hyper-parameter configuration for each PLM benchmarked on FiNER-ORD.}
\label{tb:plm_best_hyperparam_configs}
\end{table}

\section{Fine-tuning PLM Details}

\begin{table*}[]
\centering
\footnotesize
\begin{tabular}{llllll}
\hline
\textbf{Train Split} & \textbf{Test Split}& \textbf{PER}& \textbf{LOC} & \textbf{ORG} & \textbf{Weighted Avg.}\\
\hline
FiNER & FiNER & \textbf{0.9263 (0.0025)} & 0.7717 (0.0152)	& \textbf{0.6769 (0.0130)} & \textbf{0.7648 (0.0057)}  \\
CRA & FiNER & 0.5918 (0.0868) & 0.0145 (0.0020) &	0.0411 (0.0174) & 0.1730 (0.0262)  \\
CoNLL & FiNER &  0.8707 (0.0171)	& 0.7640 (0.0278) &	0.5668 (0.0678) & 0.6954 (0.0278)  \\
FiNER+CoNLL & FiNER & 0.9107 (0.0185) &	\textbf{0.7992 (0.0071)}	& 0.6441 (0.0183) & 	07522 (0.0141) \\
FiNER & CoNLL & 0.9164 (0.0299)	&0.7099 (0.0218) &	0.5622 (0.0439)	&0.7278 (0.0317)  \\
CRA & CoNLL & 0.6498 (0.0562) &	0.0492 (0.0173) &	0.2078 (0.0685) &	0.2988 (0.0316)  \\
CoNLL & CoNLL & 0.9553 (0.0080)	& 0.8974 (0.0214) & 	\textbf{0.8038 (0.0180)} & 0.8849 (0.0146)  \\
FiNER+CoNLL & CoNLL & \textbf{0.9559 (0.0047)} & \textbf{0.9018 (0.0102)} & 0.8001 (0.0965) & \textbf{0.8866 (0.0317)}
 \\
\hline
\end{tabular}
\caption{Cross-dataset and combined dataset performance analysis of FiNER, CRA, and CoNLL}
\label{tb:cross_dataset}
\end{table*}

\begin{table*}[]
\centering
\footnotesize
\begin{tabular}{llllll}
\hline
\textbf{Model} & \textbf{Test Split}& \textbf{PER}& \textbf{LOC} & \textbf{ORG} & \textbf{Weighted Avg.}\\
\hline
tner/roberta-large-conll2003 & FiNER & 0.9062 & 0.8788 & 0.8004 & 0.8474\\
roberta-large-finer & FiNER & 0.9384 & 0.8520 & 0.8352 & 0.8637 \\
 \\
\hline
\end{tabular}
\caption{Transfer learning ablation on CoNLL using \textbf{tner} framework}
\label{tb:conll2003}
\end{table*}

\label{app:best_hyperparam_configs}
No pre-training is conducted on the models before proceeding with the fine-tuning process. To determine the most suitable hyper-parameters for each model, we performed a grid search using three different learning rates (1e-4, 1e-5, 1e-6) and three different batch sizes (32, 16, 8).  We use a maximum of 100 epochs for training with early stopping criteria. If the validation F1 score doesn’t improve by more than or equal to 1e-2 in the next 7 epochs then we use the best model stored earlier as the final 6 fine-tuned model. We report the best hyper-parameter configuration for each PLM in Table \ref{tb:plm_best_hyperparam_configs}. All of our experiments are carried out using PyTorch \citep{pytorch} on an NVIDIA RTX A6000 GPU. Each model is initialized with the pre-trained version available in the Hugging Face Transformers library \citep{huggingface}. We use three different seeds (5768, 78516, 944601).

\section{Zero-shot Prompt}
\label{app:zero_shot_prompt}
"Discard all the previous instructions. Behave like you are an expert named entity identifier. Below a sentence is tokenized and each line contains a word token from the sentence. Identify "Person", "Location", and "Organisation" from them and label them. If the entity is multi token use post-fix \_B for the first label and \_I for the remaining token labels for that particular entity. The start of the separate entity should always use \_B post-fix for the label. If the token doesn't fit in any of those three categories or is not a named entity label it `Other'. Do not combine words yourself. Use a colon to separate token and label. So the format should be token:label. \{sentence\}".

\section{Comparison with CoNLL Dataset}
\label{app:combine_finer_conll}

\subsection{Combined Dataset Performance}

In order to test whether FiNER-ORD and CoNLL can be used together to achieve better NER performance, we combine the training data from both datasets. We test separately on the FiNER-ORD and CoNLL test sets. The Weighted Avg. results shown in Table \ref{tb:cross_dataset} suggest that when testing on CoNLL, FiNER-ORD can be used to complement CoNLL.

\subsection{Training on CoNLL and Testing on FiNER-ORD}
\label{app:train_CoNLL}
Section \ref{sec:dataset_comparison} highlighted how financial texts differ from general texts. In particular, financial texts tend to contain a higher ratio of organization tokens and entities compared to person and location tokens and entities. To evaluate how a model trained on the non finance specific CoNLL dataset would do on the FiNER-ORD test split, we evaluate an existing model trained on the CONNL 2003 dataset\footnote{\url{https://huggingface.co/tner/roberta-large-conll2003}}. The results in Table \ref{tb:conll2003} show that the model only performs better on LOC entities than a similar model trained on the FiNER-ORD train split. 

It is important to note that the results presented in Table \ref{tb:conll2003} are not directly comparable to those in Table \ref{tb:master_performance} due to differences in the tokenization methods and input/output formats employed by the models in each table. Specifically, the models in Table \ref{tb:master_performance} produce outputs in the form of lists of tokens and corresponding labels, whereas the models in Table \ref{tb:conll2003} generate outputs consisting of lists of entities along with their respective starting and ending positions.

\begin{table}[]
\centering
\footnotesize
\begin{tabular}{lll}
\hline
\textbf{Train Split} & \textbf{Test Split}& \textbf{Percent (\%) Overlap}\\
\hline
FiNER & FiNER & 4.9749 \\
CoNLL & FiNER & 1.1462 \\
CoNLL & CoNLL &  3.3392 \\
FiNER & CoNLL & 0.5927 \\
\hline
\end{tabular}
\caption{Cross-dataset and intra-dataset analysis of unique 4-gram lexical overlap in FiNER and CoNLL.}
\label{tb:lexical_overlap_analysis}
\end{table}

\section{Lexical Overlap Analysis}

In Table \ref{tb:lexical_overlap_analysis}, we present a cross-dataset and intra-dataset lexical overlap analysis of unique 4-grams in the train and test splits of FiNER-ORD and CoNLL. We directly follow the percent overlap calculation method presented in \citet{choubey-etal-2023-lexical}, calculating the percentage of a given test split's unique 4-grams which also exist in a given train split's set of unique 4-grams. The percent overlap between the train and test splits of FiNER is less than 5\%, so lexical similarity is relatively low within the FiNER dataset. The results in Table \ref{tb:lexical_overlap_analysis} also indicate that the FiNER test split has more lexical novelty than the CoNLL test split when compared to the CoNLL train split. Furthermore, the FiNER train split and CoNLL test split combination is most lexically different because it has the lowest percentage overlap. Thus, our lexical overlap analysis shows that FiNER is indeed lexically novel when compared to CoNLL, with respect to both the train and test splits of CoNLL, underscoring the value of our contribution with FiNER as a financial domain-specific NER dataset.

\end{document}